\pdfoutput=1

\documentclass[11pt]{article}

\usepackage{authblk}
\usepackage[]{ACL2023}

\usepackage{times}
\usepackage{latexsym}

\usepackage[T1]{fontenc}

\usepackage{algorithm}
\usepackage{algpseudocode}
\usepackage{bbm}

\usepackage[utf8]{inputenc}

\usepackage{microtype}

\usepackage{inconsolata}
\usepackage{amsmath}
\DeclareMathOperator*{\argmax}{arg\,max}

\usepackage{cleveref}
\usepackage{graphicx}
\usepackage{xcolor}
\usepackage{subcaption}
\usepackage{listings}
\usepackage{wrapfig}
\usepackage{latexsym}
\usepackage{booktabs} %
\usepackage{multirow} %
\usepackage{enumitem}

\definecolor{cblue}{HTML}{078DDD}
\definecolor{cgreen}{HTML}{8DBC70}

\newcommand{\cblue}[1]{{\color{cblue}#1}}
\newcommand{\cgreen}[1]{{\color{cgreen}#1}}

\title{Communicate to Play: Pragmatic Reasoning for Efficient Cross-Cultural Communication in Codenames}

\author[1]{Isadora White}
\author[1]{Sashrika Pandey}
\author[1]{Michelle Pan}
\affil[1]{University of California, Berkeley}

\begin{document}
\maketitle

\begin{abstract}

 Cultural differences in common ground may result in pragmatic failure and misunderstandings during communication. We develop our method Rational Speech Acts for Cross-Cultural Communication (RSA+C3) to resolve cross-cultural differences in common ground. To measure the success of our method, we study RSA+C3 in the collaborative referential game of Codenames Duet and show that our method successfully improves collaboration between simulated players of different cultures. Our contributions are threefold: (1) creating Codenames players using contrastive learning of an embedding space and LLM prompting that are aligned with human patterns of play, (2) studying culturally induced differences in common ground reflected in our trained models, and (3) demonstrating that our method RSA+C3 can ease cross-cultural communication in gameplay by inferring sociocultural context from interaction. Our code is publicly available at  \href{https://github.com/icwhite/codenames}{github.com/icwhite/codenames.}

\end{abstract}

\section{Introduction}

An English speaker from the U.K. might refer to the storage space at the back of a car as the "boot", but an English speaker from the U.S. will likely take "boot" to mean a type of shoe. The confusion that would arise in communication between these speakers is an instance of pragmatic failure \cite{thomas_cross-cultural_1983}. When humans communicate, however, they can often resolve such confusion by reasoning about the cultural background of their conversation partner, and correctly interpreting "boot" to refer to the appropriate concept. Our goal is to develop an AI system capable of pragmatic reasoning and able to adapt to new players during live interaction.

\begin{figure*}[t]
    \centering
    \includegraphics[width=\textwidth]{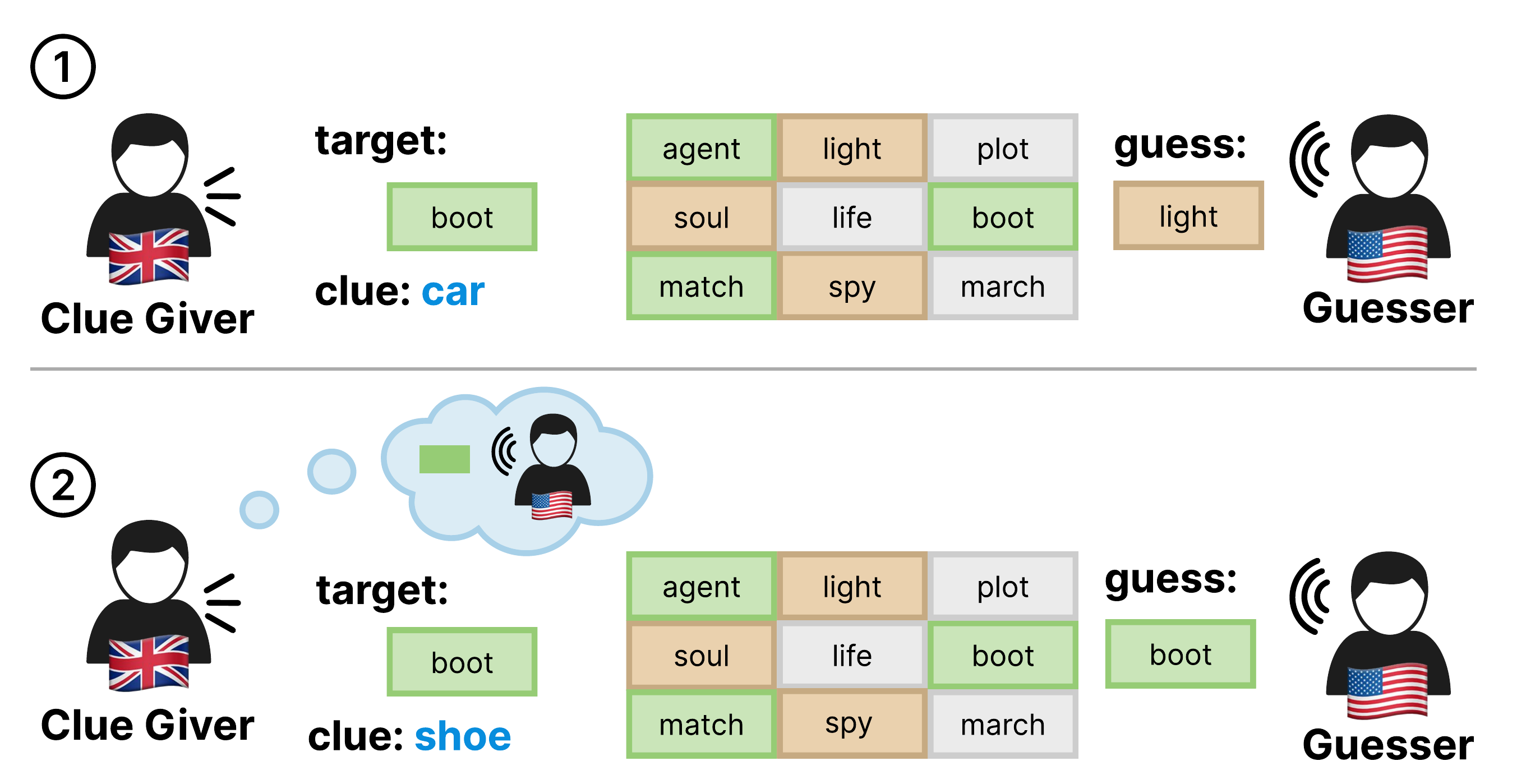}
    \caption{ \textbf{RSA+C3: Rational Speech Acts framework with Cross-Cultural Communication}. Here we model interactions in Codenames Duet between the British clue giver and the American guesser. (1) In regular gameplay, the clue giver selects a \cgreen{\bf target} and generates a \cblue{\bf clue} without considering the guesser's background. (2) Using RSA+C3, the giver considers what word the guesser may select based on their demographic background and generates a different \cblue{\bf clue} accordingly. The \color{gray} \textbf{avoid} \color{black} words will cause the game to end in an immediate loss and the \color{brown} \textbf{neutral} \color{black} words have no effect on the success or failure of the game.}
    \vspace{-1em}
    \label{fig:pull}
\end{figure*}

Existing research in cross-cultural communication focuses on single-turn interactions \cite{adilazuarda2024towards, huang2023culturally, he2024cos} or centers primarily on knowledge of cultural values or norms \cite{chiu2024culturalteaming, huang2023culturally}. However, these works miss the central aspect of inferring and adapting to socio-cultural context through interaction (e.g. an American might infer that their conversation partner is British and use this to understand what the British person means when they say "boot"). To fill this gap, we introduce our method Rational Speech Acts for Cross-Cultural Communication (RSA+C3) as illustrated in \Cref{fig:pull}. We study the effectiveness of our method by creating a test bed for culturally induced differences in common ground using the collaborative reference game Codenames Duet as described in \Cref{sec: codenames-duet}. 

First, we simulate players of Codenames Duet, using the dataset presented by \citet{cross-cultural-codes} as training data for different cultures in \Cref{sec:modeling_players}. Then, we show that these simulated players can reflect the cultural differences present in the dataset in \Cref{sec: cultural_context}. Finally, we test how well our simulated players of different cultures can play Codenames with each other \Cref{sec: interaction_section}. Through these interaction experiments, we show that our method RSA+C3 can significantly improve the win rates of games of Codenames Duet over our baseline, showing that it is inferring socio-cultural context from the interaction. Code for our experiments and to replicate our findings can be found at \href{https://github.com/icwhite/codenames}{github.com/icwhite/codenames.}

\section{Related work}
We first discuss previous work that has expanded on the Rational Speech Acts framework \cite{degen2023, GOODMAN2016818} and language games as a method of analyzing human dialogues, specifically in the context of conveying information concisely based on shared context.

\paragraph{Culture in NLP.}
State-of-the-art LLMs have been shown to struggle with multi-cultural reasoning \cite{chiu2024culturalteaming} and show uneven results across different cultures \cite{seth2024dosa}. Though prompted LLMs might reflect some understanding of cultural norms, they fail to apply reasoning to downstream inferences (e.g. inferring differences in tip culture) \cite{huang2023culturally} often producing toxic or heavily stereotyped text. Previous work has demonstrated how to personalize LLMs using prompting \cite{niszczota2023large}, influence functions \cite{he2024cos} and fine-tuning \cite{li2024culturellm}. Culturally personalized LLMs provide a useful tool for content moderation \cite{he2024cos, li2024culturellm, li2024culturepark} or sharing multi-cultural knowledge \cite{li2024culturepark}. Moreover, recent dataset and benchmark efforts \cite{fung2024massively} record a wide diversity of cultural norms. However, these papers focus mostly on norms and values (such as cultural traditions) rather than on the common ground shared between members of a culture. Norms and values refer to culturally correlated beliefs, whereas common ground refers to the assumed shared knowledge base. In contrast to the prior work, we seek to evaluate our models in their ability to infer socio-cultural differences in common ground through multi-turn interactions. 

\paragraph{Applications of RSA and Pragmatic Reasoning.} Previous work has incorporated context in the use of priors for modeling utterances via RSA, such as in using the perspective of a speaker to interpret motion verbs (e.g. "come" and "go") \cite{anderson2019} and modeling connectives in utterances (e.g. "but" and "therefore") \cite{yung-etal-2016-modelling}. RSA has also been studied as a model of human behavior through reference games, such as in differentiating ambiguous images via minimally distinguishing information \cite{frank2016}. Beyond reference games and connective utterances, RSA has been used to study discourse, particularly in the use of indirect or polite phrases \cite{lumer2022modeling}. Pragmatic reasoning plays a role in the arguments made during meetings of the UN \cite{kone2020speech}, where the ambassadors reason about the context of the others. The framework of RSA assumes that common ground is shared between parties. \Citet{degen2015wonky} adds an additional component where the probability of common ground not being shared is estimated and used to change predictions.

\paragraph{Language Games for AI.} Language games have been frequently used as a test-bed for artificial intelligence and human-AI interaction \cite{hausknecht2020interactive, ammanabrolu2022aligning, wang2022scienceworld}. Previous work explored how language models interact in realistic social environments based on choose-your-own-adventure games, finding that agents could be steered towards valuing moral requirements rather than trading them off for greater rewards \cite{pan2023machiavelli}. Codenames has been studied in the simplified format of "Codenums", which replaced words with vectors to study non-linguistic attributes of the game via a deductive agent hierarchy that tracks the internal models of other players \cite{bills2023}. Clues for the game have been generated by ranking based on document frequency and existing word embedding models \cite{koyyalagunta2021}. Sociolinguistic priors have been generated to account for the cultural context of the speaker in the simplified game "Codenames Duet" \cite{cross-cultural-codes}. We explore incorporating the speaker's sociocultural attributes across a varying set of games to explore how transferable these priors are and when this additional context could be clarifying versus superfluous.

\section{Pragmatic Reasoning with the RSA Framework and RSA+C3}

We formalize and describe the RSA framework as articulated in \citet{degen2023} and introduce our method RSA+C3. RSA formulates communication as a conversation between a listener and a speaker. For Codenames Duet, we treat the literal listener as the guesser and the pragmatic giver as the clue giver.

\subsection{RSA: Rational Speech Acts Framework} \label{sec: RSA}

In RSA, the \textit{literal listener} $L_0$ interprets meaning without considering the context. The \textit{pragmatic speaker} has the probability $P_{S_1}$ of choosing utterance $c$ given that they would like the listener to guess $g$. This is proportional to the utility $U(c, g)$ of an utterance $c$ for communicating an intended guess $g$ or in other words:
\begin{align*}
    P_{S_1}(c|g) \propto \exp( U(c, g))
\end{align*}
$U(c, g)$ represents the utility of $c$ for communicating target concepts $g$. $U$ is a trade-off between the cost of an utterance and the informativeness of $c$ defined by: 
\begin{align*}
    U(c, g) =  \ln \big( P_{L_0}(g|c) - \text{cost}(c) \big)
\end{align*}

Note that the pragmatic speaker is now selecting utterances based on the interpretations of the literal listener $P_{L_0}$. We will take the cost of the clue to be equivalent to the possibility of the guesser, or literal listener, choosing an avoid word (a word that will end the game, resulting in the other player winning) or a neutral word (a word that doesn't belong to any player's team and ends the turn without ending the game).

\subsection{RSA+C3: Rational Speech Acts for Cross-Cultural Communication} \label{sec: rsa_c3}

The RSA framework in \Cref{sec: RSA} formalizes efficient communication, but does not account for instances where common ground is not shared. We introduce RSA+C3, a method that assumes that common ground is not shared and learns to interact with an interlocutor of a different culture through live interaction. To accomplish this, we provide the RSA+C3 pragmatic speaker $S_1$ with $n$ different models representing literal listeners $L_i$ of $n$ different cultures. For each culture, we store a random variable $w_i$ where $P(w_i)$ reflects the probability that the interlocutor shares the same culture, taking inspiration from \citet{degen2015wonky}. We estimate the probability $P(w_i)$ by calculating the probability that utterance $g$ would have been chosen if the interlocutor shares the same culture and clue $c$ was given. With $g$ being the utterance observed, we then estimate:
\begin{align*}
    P(w_i) &= P_{L_i}(g|c, w_i)
\end{align*}

Then, we select a literal listener $L_i$ or guesser from the possible $n$ cultures by finding the culture that maximizes $P(w_i)$ and estimate 
\begin{align*}
    P_{S_1} (c|g) \propto \exp ( \alpha \cdot \ln( P_{L_i} (g|c) - \text{ cost }(c))) 
\end{align*}

Thereby selecting a clue $c$ to maximize informativeness to a listener belonging to a culture $i$.

\section{Task Data and Metrics}

We introduce the dataset, game, and metrics we utilize in this paper to model cross-cultural communication. 

\subsection{Codenames Duet} \label{sec: codenames-duet}
Codenames Duet is a complex referential collaborative game featuring a \textit{clue giver} and a \textit{guesser} where the clues and guesses given are based on an assumption of common ground. The board consists of 25 words, nine \textit{goal} words, three \textit{avoid} words, and 13 \textit{neutral} words. An \textit{avoid} word results in losing the game, while a \textit{neutral} has no effect. To win the game, the guesser must guess all \textit{goal} words without guessing any \textit{avoid} words. In a single turn, the \textit{clue giver} chooses a subset of the \textit{goal} words as their \textit{targets} and provide a one-word clue that the guesser uses to guess the \textit{target} words. 

\subsection{Dataset} \label{sec: dataset}
To run our experiments, we utilize Codenames Duet and the Cultural Codes \footnote{https://github.com/SALT-NLP/codenames} dataset, which contains 794 Codenames Duet games across 153 players, along with survey results containing demographic information about each player \cite{cross-cultural-codes}. The dataset is split into a train/validation/test with a 80-10-10 split and the players are different between the train and validation/test data.

\subsection{Metrics}
\label{sec: metrics}
As we use LLMs and the word embedding space to simulate interactions in Codenames, we explore our modeled givers and guessers' alignments with human data from the dataset described in \Cref{sec: dataset}.

\paragraph{Giver metrics.} In a single round, the clue giver must (1) select a set of target words from the goal words and (2) generate a clue to distinguish the intended targets from other words on the board. We define metrics for these two tasks:

\begin{itemize}
    \item \textbf{Giver target accuracy} is the proportion of the human giver's target words that are also generated by the simulated giver.
    \begin{align*}
        \frac{\text{\# giver-aligned simulated targets}}{\text{\# human giver targets}}
    \end{align*}

    \item \textbf{Clue accuracy} is the proportion of the human giver's clues that are also generated by the simulated giver.
    \begin{align*}
        \frac{\text{\# giver-aligned simulated clues}}{\text{\# human giver clues}}
    \end{align*}
\end{itemize}

We sum the number of targets and clues across multiple rounds.

\paragraph{Guesser metrics.} In a single round, the guesser selects words from the board that they believe correspond best to a given clue. We define metrics to study how well our simulated guesser aligns with both the behavior of the human guesser and the intentions of the human giver:

\begin{itemize}
    \item \textbf{Guess accuracy} is the proportion of human guesses that are also generated by the simulated guesser.
    \begin{align*}
        \frac{\text{\# guesser-aligned simulated guesses}}{\text{\# human guesser guesses}}
    \end{align*}

    \item \textbf{Guesser target accuracy} is the proportion of targets intended by the human giver that are guessed by the simulated guesser.
    \begin{align*}
        \frac{\text{\# giver-aligned simulated guesses}}{\text{\# human giver targets}}
    \end{align*}
\end{itemize}

As with the giver metrics, we sum the number of guesses and targets across rounds.

\subsection{Interactive Evaluation} \label{sec: interactive_evaluation_method}

In this work, our goal is to evaluate how simulated players of different cultures interact and collaborate to play Codenames Duet. Since Codenames Duet is a collaborative game, the main metric for whether two players are effectively communicating is the \textbf{win rate}. To ensure that a method does not increase the win rate simply by being evaluated on easier boards, we generated a fixed set of 100 boards and play a game on each board. We explain this further in \Cref{sec:interactive-eval-details}.

\begin{figure*}
    \centering
    \includegraphics[width=\textwidth]{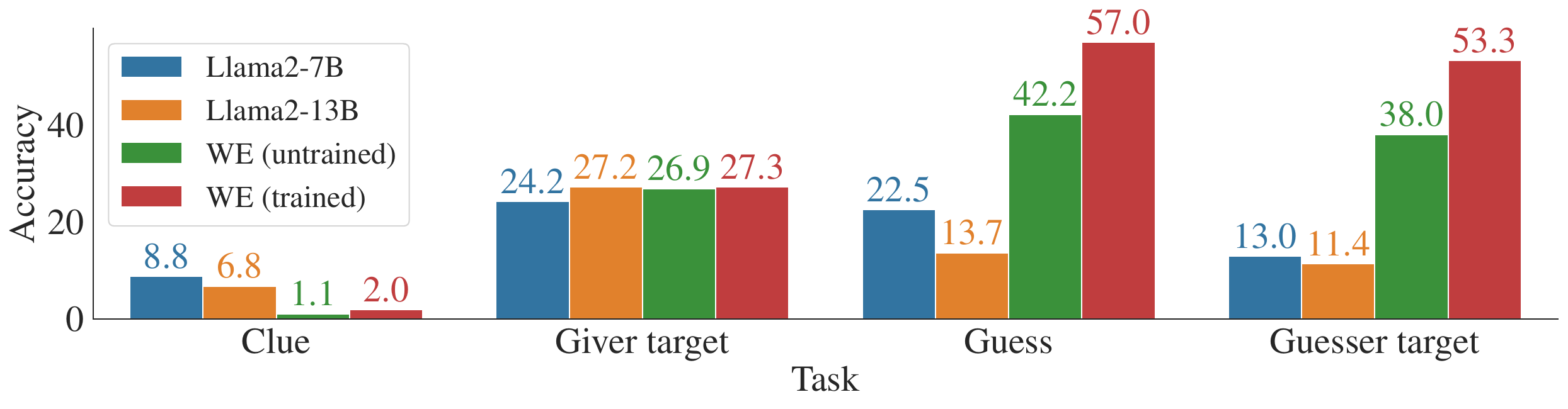}
    \caption{ \textbf{Player modeling using LLM-prompting and trained word embeddings}. The efficacy of the Llama2 chat models at simulating human players, including both the giver and guesser, varied across model size and task. Trained word embeddings consistently outperformed untrained word embeddings and generally outperformed LLM-prompting with the exception of the giver clue selection task.}
    \label{fig:llama-vs-we-all-metrics}
\vspace{-1em}
\end{figure*}

\section{Modeling Codenames Players with Word Embeddings and LLMs} 
\label{sec:modeling_players}

We explore two approaches to modeling our giver and guesser; trained word embeddings and prompting LLMs. We find that our giver and guesser based on word embeddings consistently outperform the few-shot prompted LLMs in accuracy on the human-selected guesses and targets, as illustrated in \Cref{fig:llama-vs-we-all-metrics}.

\subsection{Modelling the Guesser and Giver using Word Embeddings} \label{sec: we_guesser_giver}

The embeddings-based \textit{literal guesser} selects the most likely words based on cosine similarity between the given clue $c$ and the set of unselected words $U$. For each unselected word $u$ in $U$, the cosine similarity is given by:
\begin{align*}
    sim(c, u) &= \frac{c \cdot u}{|c| |u|}
\end{align*}

Then for the literal guesser, we estimate:
\begin{align*}
    P_{L_0}(g|c) = \frac{\exp(sim(c, g))}{\sum_{u \in U} \exp(sim(c, u))}
\end{align*}

We then select $g$ such that it maximizes $P_{L_0}(g|c)$. 
Similarly, we implement the embeddings-based \textit{literal giver} by finding the clue $c$ for target $g$ such that the similarity between $c$ and $g$ is maximized. 
\begin{align*}
    c &= \underset{c}{\argmax} \text{ } sim(c, g)
\end{align*}

Finally, we select the target concept $g$:
\begin{align*}
    g &= \underset{g}{\argmax} \text{ } \underset{c}{\argmax} \text{ } sim(c, g)
\end{align*}

\paragraph{Training Word Embeddings.}
\label{sec: train-embeddings}
To train our word embeddings we use a linear layer $f_\theta$ on top of the GloVe model \cite{pennington2014glove} and compute the embedding of a word $x$ as:
$$\text{E}(x) = f_\theta(\text{GloVe}(x))$$
During training, we aim to model the lexicon of human players by increasing the similarity between the clue and the words selected by the humans while decreasing the similarity with other words on the board.

We formalize each turn as consisting of a clue $c$, a set of available words $\{w_1, \dots, w_n\}$, and a set of selected words $S \subseteq \{1, \dots, n\}$. The training objective is then defined as: 
$$\text{loss} = -\frac{1}{|S|} \sum_{i=1}^n \log \frac{\exp(u_i)}{\sum_{j=1}^n \exp(u_j)} \mathbbm{1}\{i \in S\}$$
where $u_i$ is the cosine similarity between $w_i$ and $c$, scaled by temperature $t$:
$$u_i = \frac{\text{E}(w_i) \cdot \text{E}(c)}{|\text{E}(w_i)||\text{E}(c)|} \times \exp(t)$$

This objective is equivalent to a cross-entropy loss with equal probabilities across each selected word, and is modeled after the contrastive loss used in \citealt{radford2021learning}.

\subsection{Guesser and Giver Prompting} \label{sec: llama_prompting}

We chose to model the giver and guesser in Codenames using the Llama2 family of text and chat models \cite{touvron2023llama} due to these models being open-source.

We explore their models' accuracy across the metrics defined in \Cref{sec: metrics} with few-shot prompts. 

\paragraph{Giver.} We first query the Llama2 chat models to generate a clue using a few-shot prompt as described in \Cref{sec:clue-gen}. To allow for a diverse set of potential clues, we generated 5 clues per prompt, allowing for repeats. 
The clue giver then selects a target word for the guesser to select conditioned on the board state, as described in \Cref{sec:target-selection}.

\paragraph{Guesser.} Using a provided clue, we model the Codenames guesser by prompting a Llama2 chat model with:

\begin{lstlisting}
You are playing Codenames and are the clue guesser. You need to select one word from {all words}. Given the clue {clue}, the most likely word is 
\end{lstlisting}

We calculate the probability of a target word being generated from the list of possible target words as described in \Cref{sec:target-selection}.

\begin{figure*}
    \centering
    \includegraphics[width=\textwidth]{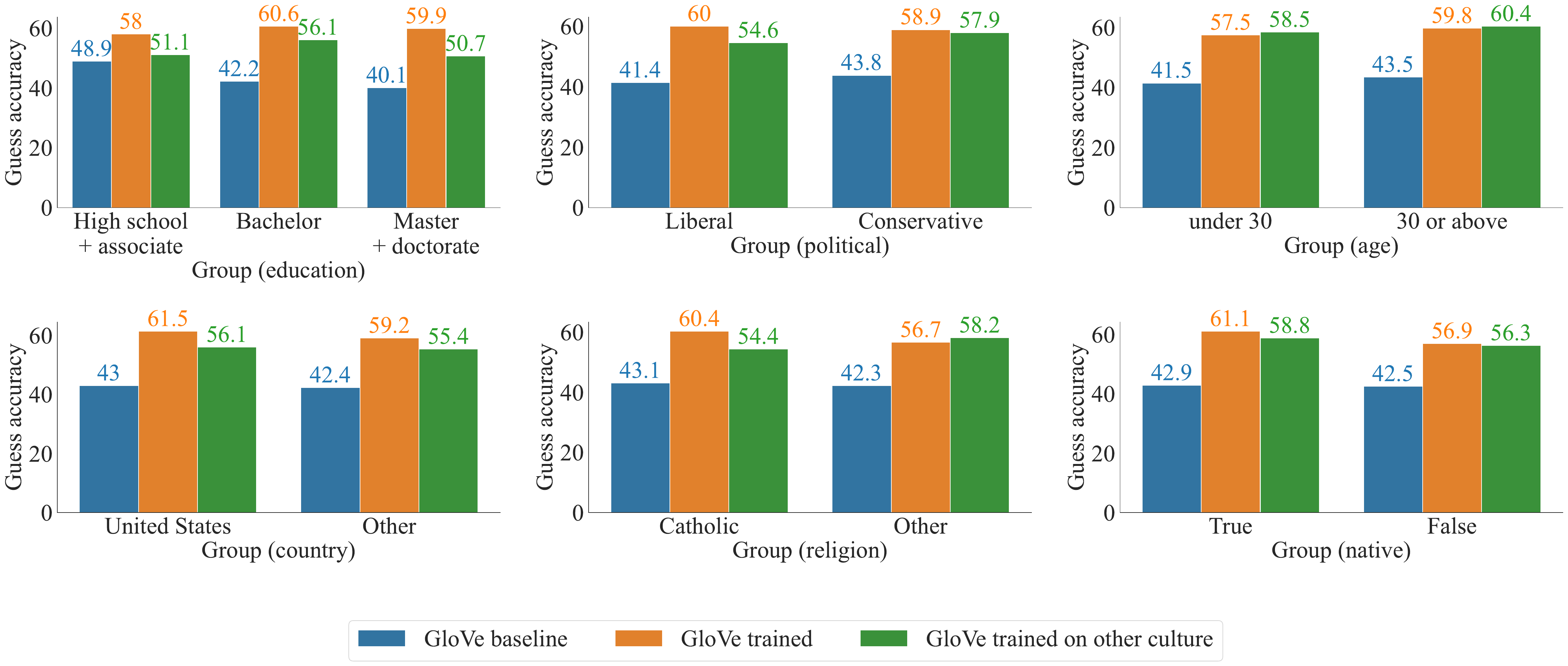}
    \caption{\textbf{Comparison of guess accuracy using embeddings trained on cultural splits against baseline GloVe and different cultural training splits.} The large difference of 9\% on the data of Master+Doctorate cultural split, between the GloVe trained on Master+Doctorate and GloVe trained on the remaining data (i.e. the difference between the orange and green bars) indicates that there are cultural patterns found in the Graduate+Bachelor data that do not occur in the remaining data. There are similar large differences in accuracy between GloVe trained on split and GloVe trained on the other split in the cultural splits on country and politics.}
    \vspace{-1em} %
    \label{fig:cultural-splits}
\end{figure*}

\section{Incorporating Cultural Context into Player Models} \label{sec: cultural_context}

To model cross-cultural communication in Codenames Duet, we must first train models to reflect the cultural background of human players. In \Cref{sec: cultural-splits}, we do this by training word embeddings using the technique described in \Cref{sec: train-embeddings} on data representing a specific demographic attribute (e.g. education). In addition, we demonstrate how few-shot prompting with cultural context can lead to higher performance, highlighting the influence of cultural priors on Codenames gameplay.

\subsection{Training embedding spaces with cultural splits} \label{sec: cultural-splits}

To model players with different cultural backgrounds, we contrastively train embeddings using the technique in \Cref{sec: train-embeddings} on subsets of the Cultural Codes dataset. We split the dataset into subsets based on various demographic and cultural attributes. We split the dataset along the axes of education (high school \& associate, bachelor, graduate), country (United States, foreign), native (true, false), political (liberal, conservative), age (under 30, over 30), and religion (Catholic, not Catholic). For some subsets of the dataset, we group the values of the cultural variables to obtain subsets with roughly equal amounts of data. We follow the procedure described in \Cref{sec:additional-embedding-results}, training for 25 epochs.

After training our embeddings, we evaluate the alignment of a literal guesser using these embeddings with the human guesses found in the hold-out validation set. The humans in the validation set are not the same humans in the training set, indicating that our predictions are extendable to other humans of a similar cultural background. Our results are displayed in \Cref{fig:cultural-splits}, with additional results in \Cref{sec:additional-embedding-results}. 

\begin{figure}
    \centering
    \includegraphics[width=0.48\textwidth]{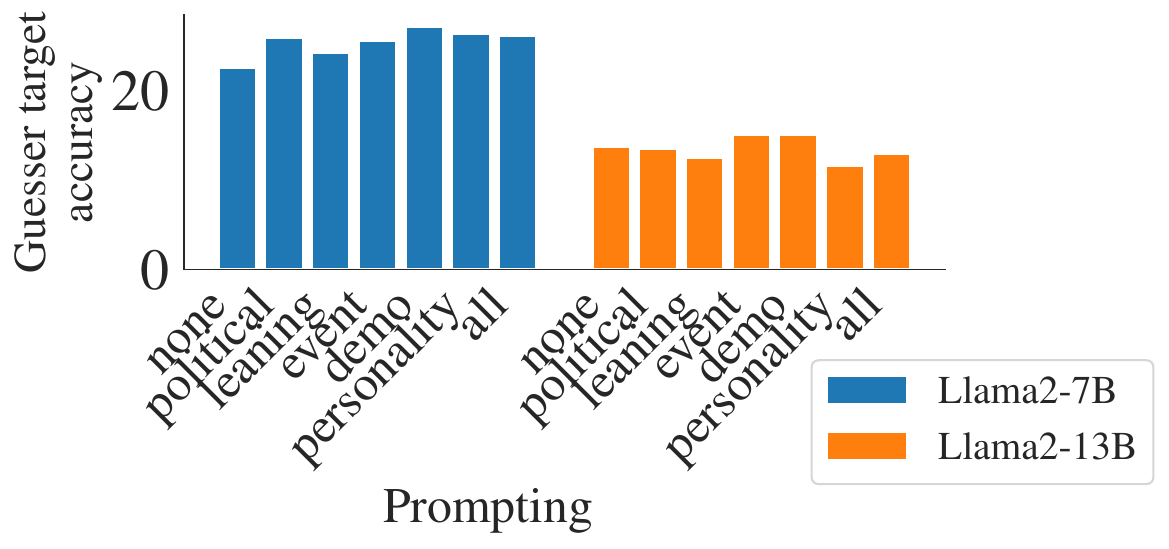}
    \caption{\textbf{Target guessing with cultural context}. Reranking potential target words based on the probabilities output by the Llama2 model simulating the clue giver and word guesser led to varying levels of guesser-aligned target word selections. Inclusion of cultural context (e.g. political leaning, personality) sometimes improved alignment with the guesser based on model size and selected demographic.}
    \label{fig:llama-guess-word}
\end{figure}

\begin{figure}
    \centering
    \includegraphics[width=0.48\textwidth]{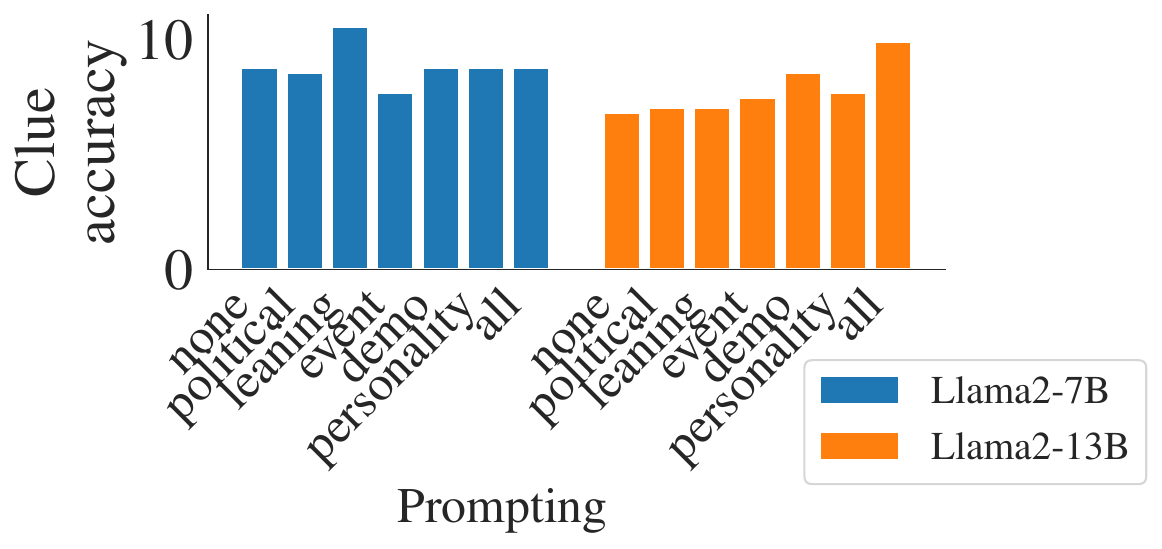}
    \caption{\textbf{Clue generation with cultural context}. Leaning notably led to an increase in accuracy for giver alignment for the 7B model while including all demographics for the 13B model led to more accurate giver-aligned generations.}
    \label{fig:llama-gen-clue}
    \vspace{-1em}
\end{figure}

\subsection{Few-shot prompting with cultural context} 

We study how different axes of demographics included in the Cultural Codes dataset could inform alignment to the human guesser and the giver, with the LLM simulating each player. In both paradigms, we prompt the openly licensed Llama2 chat models \cite{touvron2023llama} with a list of unselected words and a provided clue, asking the model to output the most likely target word. We provide information about the clue giver, as described in \Cref{sec:target-word-culture}, and study how often the model aligns to the giver and the guesser. As illustrated in \Cref{fig:llama-guess-word}, we find that including any demographic information improved alignment with the human guesser for the Llama-2-7B-Text model. Results vary for giver alignment and the 13B-Text model. Moreover, when studying the inclusion of cultural context in clue generation, we find that inclusion of all demographics increased performance in the 13B model while "leaning" (the political leaning and personality scores of the human players) increased performance for the 7B model, as shown in \Cref{fig:llama-gen-clue}. The increased performance under different cultural prompts underlines how cultural context influences the choices of the human guessers and givers in the dataset.

\section{Cross-cultural Pragmatic Reasoning in Interaction} \label{sec: interaction_section}

In \Cref{sec:modeling_players} we implemented literal listeners, and then trained literal listeners to reflect specific cultural patterns in \Cref{sec: cultural_context}. Now, we perform pragmatic reasoning with a speaker who has a \textit{different} cultural background.

\subsection{Clue Givers} \label{sec: clue_givers}

To highlight the necessity of pragmatic reasoning, we introduce our three techniques for modeling the clue giver - the literal, RSA, and RSA+C3 clue givers. 

\paragraph{Literal Clue Giver.} We evaluate the literal clue giver as described in \Cref{sec: we_guesser_giver} that selects the clue $c$ that is most similar in semantic similarity to the target $g$. 

\paragraph{RSA Clue Giver.} \label{sec: RSA_embeddings}

Recall from \Cref{sec: RSA} that we defined $P_{S_1}$ to be the probability distribution governing the actions of the pragmatic speaker. In Codenames Duet, the pragmatic speaker is the pragmatic clue giver. The clue giver must select the best clue $c$ for the target concept $g$. The cost of the clue $c$ is the probability that the guesser will instead guess avoid words $a \in A$ or neutral words $n \in N$. Therefore using $P_{L_0}$ to refer to the probability distribution of the literal guesser we use:
\begin{equation} \label{eq: rsa}
    P_{S_1} \propto \exp(\alpha \cdot (\ln P_{L_0}(g|c) - \text{cost}(c)))
\end{equation}

where 
\begin{equation} \label{eq: cost_function_codenames}
    \text{cost }(c) =  \underset{a \in A}{\max} P_{L_0}(a | c) + \delta \underset{n \in N}{\max} P_{L_0}(n |c)
\end{equation}

We introduce a neutral constant $\delta$ that governs how much to penalize the neutral words. 

\paragraph{RSA+C3 Clue Giver.} \label{sec: rsa_c3_we}

As we discuss in \Cref{sec: rsa_c3}, the RSA method described does not account for differences in common ground, or in other words, culturally introduced differences in $P_{L_0}(g|c)$. As a result, we provide $n$ word embedding models to model $n$ distributions $P_{L_i}(g|c)$. We select culture $L_i$ such that it maximizes $P(w_i)$ the posterior probability of the observed interactions if culture $i$ is shared. 
\begin{align}
    P(w_i) = P_{L_i}(g| c, w_i)
\end{align}

However, a critical component of modeling this for Codenames Duet is that there must be memory of previous interactions. Therefore $w_i$ is a smoothed average with smoothing constant $\beta$ of the estimates $P(w_i)$ after each literal guesser $L_i$ utterance. Therefore we update: 
\begin{align*}
    P(w_{i_{\text{new}}}) &= \beta \cdot P(w_{i_{\text{old}}}) + (1 - \beta) P_{L_i}(g|c, w_i)
\end{align*}

We then estimate $P_{S_1}$ the same way as in $\Cref{eq: rsa}$ but using $P_{L_i}$ so:
\begin{align*}
    P_{S_1}(c|g) \propto \exp(\alpha \cdot (\ln P_{L_i}(g|c) - \text{cost}(c)))
\end{align*}

Then we select our clue via:
\begin{align*}
    c = \underset{c}{\argmax} P_{S_1}(c|g) 
\end{align*}

\subsection{Interactive Evaluation Results} \label{sec:interactive-evaluation}
\begin{figure*}
    \centering
    \begin{subfigure}[b]{0.45\textwidth}
        \centering
        \includegraphics[width=\textwidth]{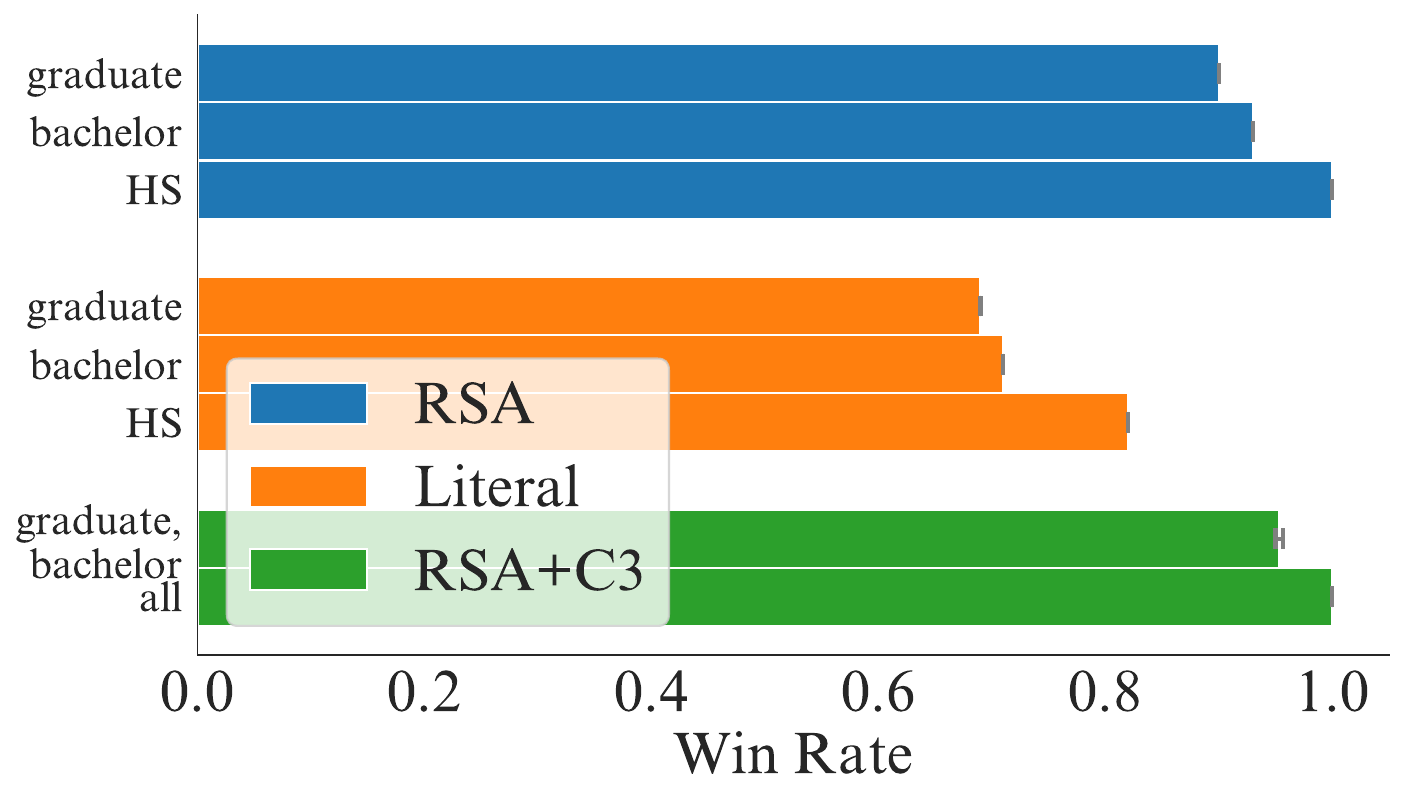}
        \caption{Word Embedding (High School) Guesser}
        \label{fig:interactive_eval_we_we}
    \end{subfigure}
    \hfill
    \begin{subfigure}[b]{0.45\textwidth}
        \centering
        \includegraphics[width=\textwidth]{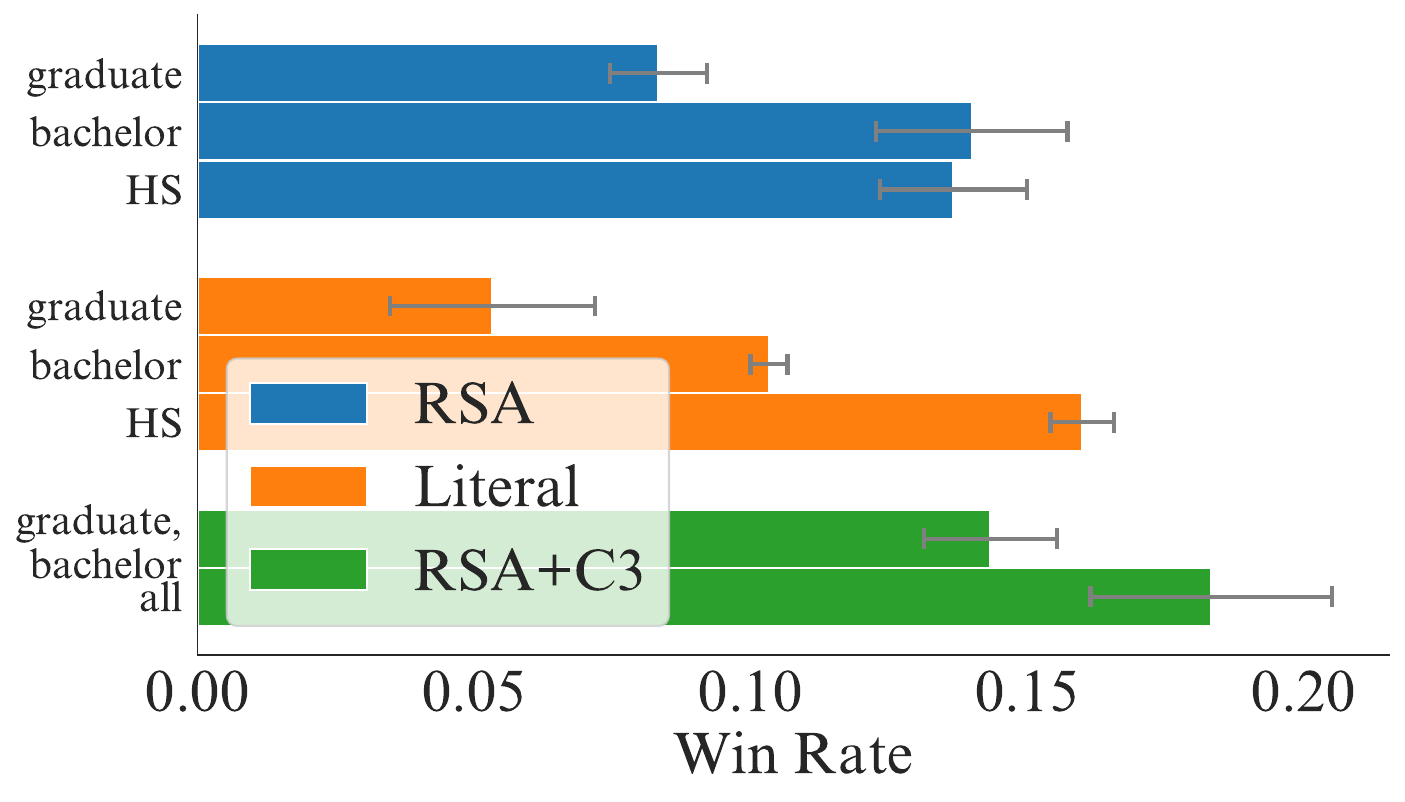}
        \caption{Llama2-Text-7B Guesser}
        \label{fig:interactive_eval_llama_we}
    \end{subfigure}
    \caption{\textbf{Interactive Evaluation across RSA, Literal, and RSA+C3 Guessers.} We evaluate RSA, Literal, and RSA+C3 givers across guessers simulated by word embedding training and LLM prompting. In \Cref{fig:interactive_eval_we_we}, we study interactions with a word embeddings guesser trained on data belonging to players whose highest level of education completed was high school. The "graduate, bachelor" RSA+C3 giver, initialized on both cultural backgrounds, achieved the highest win rate, greater than RSA givers initialized on either "graduate" or "bachelor" alone. We used an LLM-prompted guesser in \Cref{fig:interactive_eval_llama_we} and found that the RSA+C3 giver initialized with all provided education options ("graduate, bachelor, HS") achieved the highest win rate, outperforming all RSA and Literal givers. To select the most appropriate neutral penalty of $0.1$ and $\alpha$ as $0.5$ we perform hyperparameter tuning as described in \Cref{sec: hyperparameter_tuning}. To calculate error bars we do three runs and take the standard error mean. }
    \label{fig:education-interactive-results}
    \vspace{-1em}
\end{figure*}

As described in \Cref{sec: interactive_evaluation_method}, we evaluate the performance of two players of different cultures during interaction. To do this, we select the demographic in the dataset such that simulated players have the largest cultural difference as observed in \Cref{fig:cultural-splits} - education. 

We evaluate our literal, RSA, and RSA+C3 clue givers against two different guessers: a guesser trained to reflect a player with a high school or associates degree and Llama2-7B-Chat prompted as described in \Cref{sec: llama_prompting}. We evaluate with the Llama2-7B-Chat guesser to simulate an unknown culture that the clue giver must adapt to. To ensure that players reflect different cultures we evaluate simulated players with a graduate or undergraduate degree when playing against the player with a high school degree.

While the inclusion of the traditional RSA framework leads to significant improvements in contrast to the literal giver, our results in \Cref{fig:education-interactive-results} demonstrate that including pragmatic reasoning and cross-cultural communication via RSA+C3 leads to a greater win rate regardless of whether the guesser is trained word embeddings or a prompted LLM.

\section{Discussion}
Using Codenames Duet as a testbed for studying cross-cultural communication, we demonstrated that our simulated players are capable of reflecting human gameplay and their sociocultural patterns. We utilize our player models reflecting different sociocultural backgrounds to emulate pragmatic failure in live gameplay. This enables us and future researchers to measure the collaborative ability between agents of different backgrounds - if the win rate of Codenames Duet is higher, then the difference in common ground is more easily overcome.

As the full complexity of cross-cultural communication cannot only be captured through Codenames Duet, directions for future work include applying these techniques to more complex utterances with more nuanced cultural differences and studying the resulting interactive gameplay. 

Overall, we find that introducing cultural context as a way for givers and guessers to communicate in Codenames Duet gameplay increases alignment with human data based on the subset of culture involved. Our results across various methods of simulating players and different cross-sections of demographics demonstrate the significance of continuing to study the impact of cultural context in speaker and listener communication.

\section{Limitations}

In our paper, we train models to reflect various cultural attributes as shown in \Cref{fig:cultural-splits} and evaluate our method RSA+C3 to resolve pragmatic failure due to cultural differences such as education level in \Cref{fig:education-interactive-results}. However, the cultures are not equally represented in the cross-cultural codes dataset \cite{cross-cultural-codes} we used with the participants being majority White (78\%) and liberal (58\%). Therefore some cultural differences are not as pronounced as they would be in a more balanced dataset. We encourage future work to study gameplay on diverse data and explore communication in gameplay on a broader range of cultural subsets.

\section{Broader impacts statement}
While cultural context can be a useful tool in informing clue generation and target selection in games like Codenames, we caution against leaning heavily on these demographics due to the potential for stereotype-based associations. Previous work has demonstrated the propensity for language models to incorporate biases into generations \citep{bias_gender_llm}. Although we are interested in seeing future work explore how culture can inform communication, allowing for both speakers and listeners to update their mental models of the other conversational participant, we acknowledge that leaning too heavily on these demographics can lead to potentially harmful assumptions.

\section{Acknowledgements}
We would like to thank Alane Suhr and Lianhui Qin for their guidance and feedback on the paper. Additionally, thanks to Jakub Grudzien Kuba for his help conducting data analysis experiments in initial versions of this work.

\bibliographystyle{acl_natbib}
\bibliography{custom}

\appendix
\section{Experiment details for simulating givers and guessers using LLMs}
\label{sec:llama_exps}

Here we elaborate on the framework for our experiments in clue and target selection using the Llama2 family of LLMs, as described in \Cref{sec:modeling_players}. We chose to use Llama2 because it is open-source and was the most recent family of Llama models available at the time.

For all of the following experiments, we used default hyperparameters as provided in the open-source Llama2 code \footnote{https://github.com/meta-llama/llama} and model sizes of 7B and 13B. The following experiments were conducted over the validation set of the Cultural Codes dataset.

\subsection{Clue generation}
\label{sec:clue-gen}
We prompted the 7B and 13B Llama2-Chat models to generate clues using the following few-shot prompt, allowing for a flexible free-form text generation informed by prior examples of a Codenames-style clue:

\begin{lstlisting}
You are playing Codenames. You can only give clues which are one word. One clue will apply to multiple targets. Words to avoid are {avoid words}. Neutral words are {neutral words}. For the group of target words ['fall', 'spring', and 'leaf'] the best clue is 'season'. For the group of target words ['round', 'cylinder'] the best clue is 'circle'. For the target words {target words} the best clue is '
\end{lstlisting}

The target words were preselected from the Cultural Context dataset, allowing us to study the LLM's alignment with a human clue giver.

\subsection{Target selection}
\label{sec:target-selection}
Using the Llama2 Text models, we used the following prompt to extract potential target words. 

\begin{lstlisting} 
You are playing Codenames and need to select a target word for your partner to guess. Words to avoid are {avoid words}. Neutral words are {neutral words}. Goal words are {goal words}. The best target word for your partner to guess is '
\end{lstlisting}

As the game is constrained to selecting target words from the set of goal words, we calculated the probability of the model generating each of the goal words as the completion to the prompt, then identified the most probable generations as the selected target words.

\subsection{Target word selection under cultural context}
\label{sec:target-word-culture}
We prompted the Llama2 Text models with the following prompt, optionally including the giver's demographics. Similar to our experiment with target selection in \Cref{sec:target-selection}, we selected the generation under the set of possible target words (i.e. restricted to the set of goal words) that had the highest probability. 

\begin{lstlisting}
You are playing Codenames. The possible words are {words}. Here is some information about the clue giver: {cultural context}. For the hint {clue}, the most likely target word is 
\end{lstlisting}

As demographics were verbose, we provided them as a comma-separated list of values. For example, one possible prompt addition could be:

\begin{lstlisting}
Here is some information about the clue giver: age: 29, gender: female, country: united states, native: true.
\end{lstlisting}

The demographics we used in \Cref{fig:llama-guess-word} consist of the demographic questions in the Cultural Codes dataset in Appendix D.2 of \citet{cross-cultural-codes}. We additionally extracted the political context from the broader political leaning category (abbreviated in the figure as ``leaning").

\begin{figure*}
    \centering
    \includegraphics[width=\textwidth]{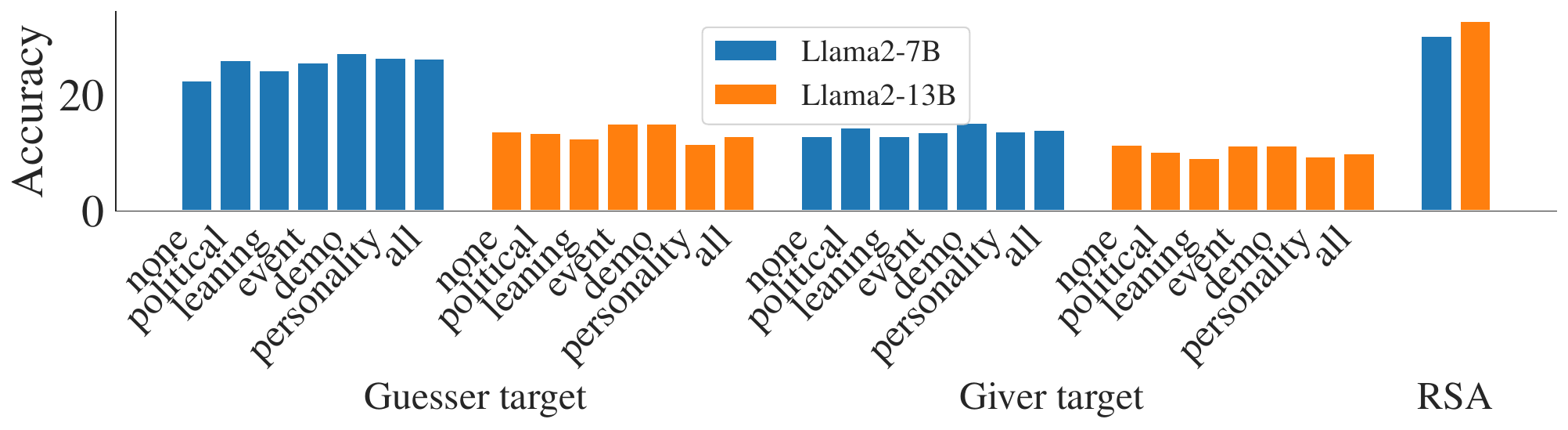}
    \caption{\small \textbf{Giver and guesser alignment for target selection}. RSA resulted in greater accuracy across both model sizes while model effectiveness varied across the cultural demographic that was included. Definitions of each cultural split can be found in Appendix D.2 of \citet{cross-cultural-codes}.}
    \label{fig:llama-giver-guesser-align}
\end{figure*}

Notably, we calculated accuracy for giver alignment versus guesser alignment with separate target words. Alignment with the giver meant selecting target words that were intended by the human giver for the guesser to select. Alignment with the guesser meant selecting target words that the human guesser selected given a similar set of information as provided in the prompt above, regardless of the giver's original intentions. As multiple target words could be selected per round, we computed the accuracy as the total number of correct target words divided by the total number of intended target words. Full results for both giver and guesser alignment can be found in \Cref{fig:llama-giver-guesser-align}.

\subsection{Clue generation under cultural context}
We iterated on our clue generation experiments from \Cref{sec:clue-gen} by using a similar approach to \Cref{sec:target-word-culture}, drawing pre-specified demographics for the guesser to inform the giver's clues. We generated prompts of the following format:

\begin{lstlisting}
You are playing Codenames. You can only give clues which are one word. One clue will apply to multiple targets. Words to avoid are {avoid words}. Neutral words are {neutral words}. Here is some information about the clue guesser: {cultural context}. For the group of target words ['fall', 'spring', and 'leaf'] the best clue is 'season'. For the group of target words ['round', 'cylinder'] the best clue is 'circle'. For the target words {target words} the best clue is '
\end{lstlisting}

\subsection{Rational speech acts framework}
\label{sec:rsa-details}
In our extension of the RSA framework, we first queried the Llama2 chat models to generate a clue using the same clue generation prompt from \Cref{sec:clue-gen}. To allow for a diverse set of potential clues, we generated 5 clues per prompt, allowing for repeat clues.

Using these clues, we then queried the model to select a target word using the following prompt:

\begin{lstlisting}
You are playing Codenames and are the clue guesser. You need to select one word from {all words}. Given the clue {clue}, the most likely word is 
\end{lstlisting}

We calculated the probability of a target word being generated from the list of possible target words as described in \Cref{sec:target-selection}. Following both queries, we calculated the probability of the guesser's target word generation under a given clue as the sum of the individual probabilities of the target word being generated by the LlamaGuesser and the clue being generated by the LlamaGiver. Comparing these cumulative probabilities across all target word and clue pairs allowed us to \emph{rerank} the probability of a given utterance. 

As every prompt in the Cultural Codes dataset had the human giver's intended target words (sometimes multiple), we selected the top unique target words and calculated the accuracy of our LlamaGiver and LlamaGuesser together. Here, accuracy is based on alignment with the human giver. For clue selection, we selected the corresponding clue paired with the most probable target word.

\section{Additional embedding training results}
\label{sec:additional-embedding-results}

\subsection{Target accuracy}

\begin{figure*}
    \centering
    \includegraphics[width=\textwidth]{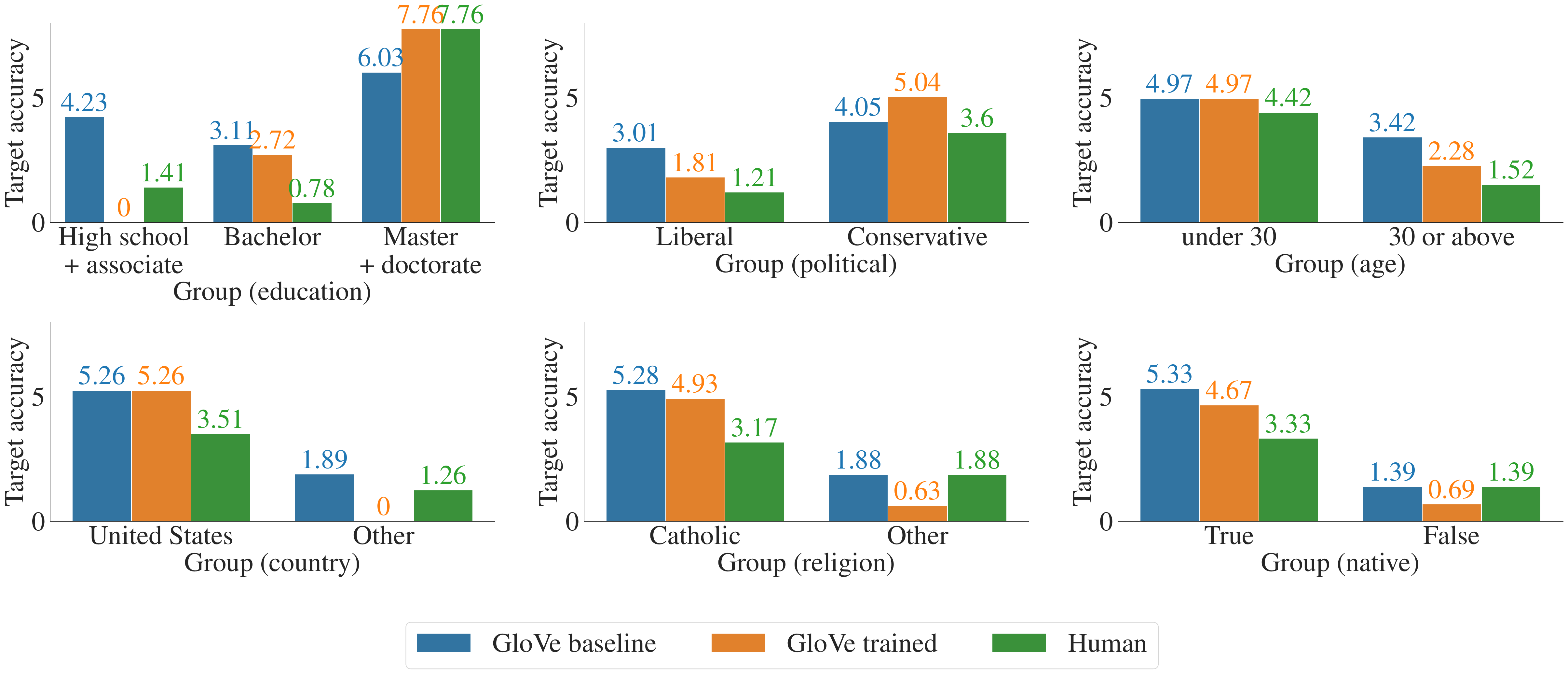}
    \caption{Comparison of target accuracy using embeddings trained on cultural splits against baseline GloVe embeddings. Target accuracy measures the performance of embeddings in correctly selecting the intended target words chosen by the clue giver. In green is the performance of the human guessers in the dataset.}
    \label{fig:embedding-target-acc}
\end{figure*}

We evaluate the performance of trained embeddings in selecting correct targets, with results shown in \Cref{fig:embedding-target-acc}. Our method for training embeddings generally does not result in improved target accuracy. In fact, since the untrained GloVe embeddings perform better than human guessers in selecting the intended targets, training on human data decreases the target accuracy in many cases.

\subsection{Improvement over baselines}

We include our numerical results in Tables \ref{embedding-guess-acc}, \ref{embedding-training-split-comparison}, \& \ref{embedding-target-acc}, showing accuracy of trained embeddings compared to that of baselines.

\begin{table*}
    \centering
    \begin{tabular}{llll} \toprule
    Group & 
    \multicolumn{1}{p{2.5cm}}{\raggedright GloVe baseline guess acc.} & 
    \multicolumn{1}{p{2.5cm}}{\raggedright GloVe trained guess acc.} & 
    \multicolumn{1}{p{2.5cm}}{\raggedright \% improvement} \\ \midrule
    \multicolumn{1}{p{4cm}}{\raggedright Education: high school, associate}   & 48.86     & 57.95   & 49.13 \\
    Education: bachelor                 & 42.21	    & 60.55   & 18.6  \\
    Education: graduate                 & 40.14	    & 59.86   & 40.16 \\
    Gender: female                      & 38.97	    & 56.34   & 45.07 \\
    Gender: male                        & 45.42	    & 63.09   & 43.03 \\
    Country: united states              & 42.99	    & 61.49   & 38.90 \\
    Country: foreign                    & 42.39	    & 59.24   & 43.45 \\
    Native: true                        & 42.90     & 61.08   & 39.75 \\
    Native: false                       & 42.51	    & 56.89   & 42.38 \\
    Political: liberal                  & 41.36	    & 60.00   & 34.35 \\
    Political: conservative             & 43.81	    & 58.86   & 33.83 \\
    Age: under 30                       & 41.49	    & 57.45   & 57.45 \\
    Age: over 30                        & 43.50     & 59.82   & 59.82 \\
    Religion: catholic                  & 43.08	    & 60.38   & 60.38 \\
    Religion: not catholic              & 42.29	    & 56.72   & 56.72 \\
    All                                 & 43.16	    & 60.50   & 40.18 \\ \bottomrule
    \end{tabular}
    \caption{\label{embedding-guess-acc}
    Guess accuracy of trained embeddings across dataset splits before and after training with our contrastive learning algorithm described in 
    }
\end{table*}

\begin{table*}
    \centering
    \begin{tabular}{lllll} \toprule
    Group & 
    \multicolumn{1}{p{2cm}}{\raggedright Same split guess acc.} & 
    \multicolumn{1}{p{2.5cm}}{\raggedright Other split guess acc.} & 
    \multicolumn{1}{p{2.5cm}}{\raggedright \% difference between cultures} \\ \midrule
    \multicolumn{1}{p{4cm}}{\raggedright Education: high school, associate} & 57.95 & 51.14 & 13.32 \\
    Education: bachelor                 & 60.55     & 56.06     & 8.01  \\
    Education: graduate                 & 59.86     & 50.70     & 18.07 \\
    Gender: female                      & 56.34     & 56.81     & ---   \\
    Gender: male                        & 63.09     & 58.50     & 7.85  \\
    Country: united states              & 61.49     & 56.12     & 9.57  \\
    Country: foreign                    & 59.24     & 55.43     & 6.87  \\
    Native: true                        & 61.08     & 58.81     & 3.86  \\
    Native: false                       & 56.89     & 56.29     & 1.07  \\
    Political: liberal                  & 60.00     & 54.55     & 9.99  \\
    Political: conservative             & 58.86     & 57.86     & 1.73  \\
    Age: under 30                       & 57.45     & 58.51     & ---   \\
    Age: over 30                        & 59.82     & 60.42     & ---   \\
    Religion: catholic                  & 60.38     & 54.40     & 10.99 \\
    Religion: not catholic              & 56.72     & 58.21     & ---   \\ \bottomrule
    \end{tabular}
    \caption{\label{embedding-training-split-comparison}
    Comparison of guess accuracy when embeddings are trained on data from the same culture vs. data from different cultures. 
    }
\end{table*}

\begin{table*}
    \centering
    \begin{tabular}{lllll} \toprule
    Group & 
    \multicolumn{1}{p{2cm}}{\raggedright Human target acc.} & 
    \multicolumn{1}{p{2.5cm}}{\raggedright GloVe baseline guess acc.} & 
    \multicolumn{1}{p{2.5cm}}{\raggedright GloVe trained guess acc.} & 
    \multicolumn{1}{p{2.5cm}}{\raggedright \% improvement} \\ \midrule
    \multicolumn{1}{p{4cm}}{\raggedright Education: high school, associate}   & 1.41	& 4.23 & 0.00 & ---   \\
    Education: bachelor                 & 7.78      & 3.11      & 2.72      & ---   \\
    Education: graduate                 & 7.76      & 6.03      & 7.76      & 28.6   \\
    Gender: female                      & 1.12      & 4.47      & 2.80      & ---   \\
    Gender: male                        & 3.77      & 3.77      & 3.77      & 0.00   \\
    Country: united states              & 3.51      & 5.26      & 5.26      & 0.00   \\
    Country: foreign                    & 1.26      & 1.89      & 0.00      & ---   \\
    Native: true                        & 3.33      & 5.33      & 4.67      & ---   \\
    Native: false                       & 1.39      & 1.39      & 0.69      & ---   \\
    Political: liberal                  & 1.21      & 3.01      & 1.81      & ---   \\
    Political: conservative             & 3.60      & 4.05      & 5.04      & 24.22 \\
    Age: under 30                       & 4.42      & 4.97      & 4.97      & 0.00  \\
    Age: over 30                        & 1.52      & 3.42      & 2.28      & ---   \\
    Religion: catholic                  & 3.17      & 5.28      & 4.93      & ---   \\
    Religion: not catholic              & 1.88      & 1.88      & 0.63      & ---   \\
    All                                 & 2.70      & 4.05      & 3.60      & ---   \\ \bottomrule
    \end{tabular}
    \caption{\label{embedding-target-acc}
    Target accuracy of trained embeddings across dataset splits.
    }
\end{table*}

\section{RSA Extensions}

In a dialogue, there is both a \textit{speaker} and a \textit{listener}. The goal of the \textit{speaker} is to communicate concepts that the \textit{listener} aims to interpret. The standard RSA framework assumes that the speaker and listener share common ground \cite{degen2023}. In cross-cultural communication, this assumption is false. We propose a method for modeling the repair process \cite{pickering2004toward} of two speakers aiming to find common ground. 

In RSA formulations, the (abstract) \textit{literal listener} $L_0$ interprets meaning based on literal semantics. The \textit{pragmatic speaker} $S_1$ reasons about the literal listener and chooses utterances to optimize informativeness while minimizing the cost (e.g. length). 
Formally, let $w$ represent an abstract variable referred to as \textit{world} in \citet{degen2023} and $m$ stand for the meaning that the speaker wants to convey with their utterance $u$. Importantly, $w$ can be instantiated by different situations or contexts in which the interlocutors find themselves. The joint probability distribution of these variables, conditioned on $w$, factorizes as 
\begin{align}
    \label{eq:speaker}
    P(m, u| w) = P(m|w)P_{S_1}(u|w, m), 
\end{align}
where $P_{S_1}$ is governed by speaker $S_1$. 
The goal of pragmatic listener $L_1$ is to comprehend the meaning $m$ and infer meaning $m$ given $w$ and $S_1$'s utterance $u$. Using Bayes' rule, this probability is proportional to: 
\begin{align}
    \label{eq:listener}
    P_{L_1}(m|w, u) \propto P(m|w)P_{L_1}(u|w, m).
\end{align}
The subtle assumption made by this equation is that the probability over meanings, given world, is independent of the interlocutor, and thus $L_1$ reasons about it the same way the speaker does.
We believe that this is \textsl{not true}. 
The response, and therefore a meaning to communicate, to a situation depends tightly on the speaker, and can be shaped by factors such as cultural or demographic background.
Hence, in the context of cross-cultural communication, \Cref{eq:speaker} should be written as: 
\begin{align}
    P(m, u| w) = P_{\color{blue}S_1\color{black}}(m|w)P_{S_1}(u|w, m), \nonumber
\end{align}
and \Cref{eq:listener} would read:
\begin{align}
    P_{L_1}(m|w, u) \propto P_{\color{blue}L_1\color{black}}(m|w)P_{L_1}(u|w, m).\nonumber
\end{align}

In this paper, we will model two different \textit{literal listeners} and respective \textit{pragmatic speakers} with overlapping but not identical prior beliefs. We will model the different literal listeners and pragmatic speakers using prompting and/or training. Therefore these pragmatic speakers will have different subjective prior beliefs, reflecting the scenario of cross-cultural communication. We then seek to learn a \textit{pragmatic listener} with incorrect or without access to the prior beliefs of the \textit{pragmatic speaker}. 
\begin{align*}
    P_{L_1}(m, w|u) = P_{S_1}(u|m, w) \cdot P(m|w) \cdot P(w)
\end{align*}

Where the variable captures whether the world is normal or wonky such that:
\begin{align*}
    P(m|w) \propto \begin{cases}
        P_{usual} (m) & \text{if not } w,\\
        P_{backoff}(m) & \text{ if } w
    \end{cases}
\end{align*}

In this case, $P_{usual}$ is the prior probability in the scenario where the world is "normal" and $P_{backoff}$ is the prior probability where the world is "wonky". This backoff probability is a uniform distribution. The value of $w$ is inferred from the utterances $u$ of the pragmatic speaker $S_1$ by the pragmatic listener $L_1$ based on how unlikely the utterances $u$ are in the context of the pragmatic listener's prior beliefs. To calculate the posterior beliefs of the pragmatic listener about the meaning $w$:
\begin{align*}
    P_{L_1}(m | w) \propto \sum_{w} P_{L_1}(m, w|u)
\end{align*}

The pragmatic listener's posterior probabilities are a mixture of the computation and a backoff prior based on how likely it is that $w$ is true and the world is "wonky". In cross-cultural communication, the "wonky" world represents the case where the assumed common ground does not exist or is different in some way. In this paper, we hypothesize that RSA and the concept of wonky world can assist in understanding cross-cultural communication in the context of Codenames Duet and predict when common ground is not held between agents. 

\section{Hyperparameter Tuning for RSA and RSA+C3} \label{sec: hyperparameter_tuning}

\begin{figure}
    \centering
    \includegraphics[width=0.5\textwidth]{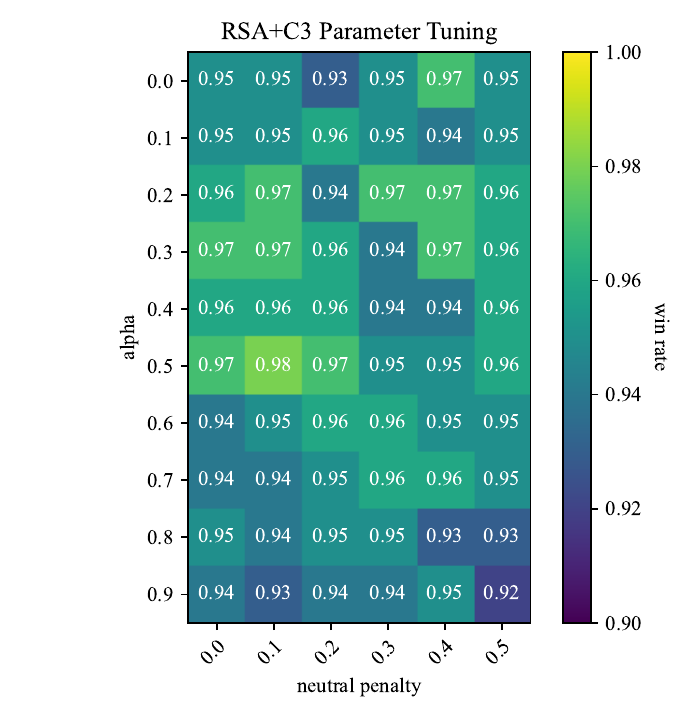}
    \caption{\textbf{Hyperparameter Tuning for RSA+C3 across the axes of alpha and neutral penalty}. We find that a neutral penalty of 0.1 and an alpha of 0.5 performed the best. 
    \label{fig:rsa_c3_hyperparameter_1}}
\end{figure}

\begin{figure}
    \centering
    \includegraphics[width=0.5\textwidth]{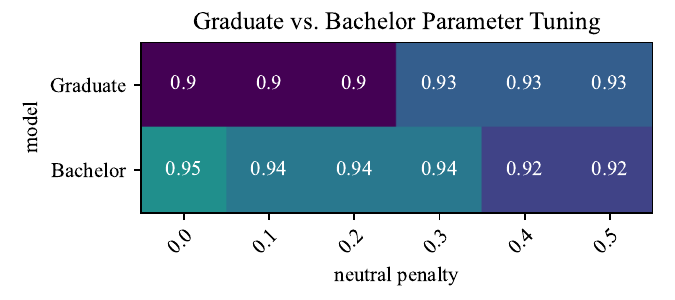}
    \caption{\textbf{Hyperparameter Tuning for RSA+C3 across the axes of alpha and neutral penalty}. We find that a neutral penalty of 0.1 or 0.3 performed the best across the different cultures. 
    \label{fig:rsa_c3_hyperparameter_2}}
\end{figure}

In this section, we tune the hyperparameters for RSA+C3 and RSA methods. We find that many of the hyperparameters perform similarly but the best performance is achieved with a neutral penalty of $0.1$ and an alpha of $0.5$. We include our tuning findings in \Cref{fig:rsa_c3_hyperparameter_1} and \Cref{fig:rsa_c3_hyperparameter_2}

For RSA, there were not significant differences observed for the different values of the neutral penalty. 

\section{Interactive Evaluation Experiments}
\label{sec:interactive-eval-details}

We run experiments with 1 target, because of higher win rates. We ran the experiments for Llama2-7B-Text for 100 games and the one for the High School guesser for 1000 games. We ran less games under Llama due to time restrictions.

To make sure that the games all occur on the same set of boards, we generate a fixed set of boards to be used for each experiment. We do this by generating a set of $n$ board each with a unique seed and hold the seeds constant. This allows us to easily scale up a number of boards while ensuring that the boards are the same for each run and each experiment. 

\section{Qualitative examples of cultural context}
\label{sec:qualitative-culture}

Below are qualitative examples demonstrating miscommunications between two simulated players initialized with different cultural backgrounds; from experiments on education backgrounds in \Cref{sec: cultural-splits}, the following interactions are between a graduate giver and a high school guesser which can be resolved by RSA+C3. 

This example shows a graduate giver thinking of “chemical compound” instead of a chemical as a poison as the high school guesser inferred. 

\begin{lstlisting}
CLUE GIVER'S TURN
Targets selected: compound
Clue: chemical

GUESSER'S TURN
Guessed words: poison
Result: avoid
\end{lstlisting}

This example shows a graduate giver highlighting an association of programming with coding rather than a degree program as the high school guesser inferred. 

\begin{lstlisting}
CLUE GIVER'S TURN
Targets selected: code
Clue: program

GUESSER'S TURN
Guessed words: degree
Result: avoid
\end{lstlisting}

Note that here we are using the education as the main distinguishing factor of culture, which would define which concepts are most topical for a given user.

\end{document}